# How to Learn a Model Checker

Dung Phan
Department of Computer Science,
Stony Brook University
Stony Brook, New York, USA

Radu Grosu
Cyber-Physical Systems Group,
Technische Universitat Wien
Vienna, Austria

Nicola Paoletti
Department of Computer Science,
Stony Brook University
Stony Brook, New York, USA

Scott A. Smolka
Department of Computer Science,
Stony Brook University
Stony Brook, New York, USA

Scott D. Stoller
Department of Computer Science,
Stony Brook University
Stony Brook, New York, USA

## ABSTRACT

We show how machine-learning techniques, particularly neural networks, offer a very effective and highly efficient solution to the approximate model-checking problem for continuous and hybrid systems, a solution where the general-purpose model checker is replaced by a model-specific classifier trained by sampling model trajectories. To the best of our knowledge, we are the first to establish this link from machine learning to model checking. Our method comprises a pipeline of analysis techniques for estimating and obtaining statistical guarantees on the classifier's prediction performance, as well as tuning techniques to improve such performance. Our experimental evaluation considers the time-bounded reachability problem for three well-established benchmarks in the hybrid systems community. On these examples, we achieve an accuracy of 99.82% to 100% and a false-negative rate (incorrectly predicting that unsafe states are not reachable from a given state) of 0.0007 to 0. We believe that this level of accuracy is acceptable in many practical applications and we show how the approximate model checker can be made more conservative by tuning the classifier through further training and selection of the classification threshold.

## 1 INTRODUCTION

The formal verification community has taken note of the ongoing improvements to and increasing applications of machine learning (ML). In particular, model checking (MC) techniques have been applied to the safety verification of state-of-the-art ML technology, including Deep Neural Networks [31, 41, 45]. To the best of our knowledge, however, no one has considered the inverse problem: How can ML techniques be applied to the MC problem? Phrasing this another way: *How can one train a neural network for MC purposes?*

This is the problem we consider in this paper. Specifically, we show how it is possible to train a neural network (NN) for the purpose of model checking continuous and hybrid systems (HSs). Given an HS $\mathcal{M}$ with state-space $S$, a state $s \in S$, a time bound $T$, a set of "unsafe" states $U \subset S$ (or states of interest for any reason), we consider the *time-bounded reachability* problem for HSs: is it possible for $\mathcal{M}$, starting in $s$, to reach a state in $U$ within time bound $T$? As such, the NN we obtain is a *classifier* $f$ of the form $f : S \rightarrow \{\text{false}, \text{true}\}$, where a negative classification ($f(s) = \text{false}$) means that a state in $U$ cannot be reached from $s$ within time $T$, and a positive classification ($f(s) = \text{true}$) means a state in $U$ can be reached from $s$ within time $T$.

A classifier of this type is subject to *false positives* (a state $s$ is deemed positive when it is actually negative) and, more importantly, *false negatives* ($s$ is deemed negative when it is actually positive). We show that the false-negative rate can be improved by adapting the NN on counterexamples identified during additional training.

We refer to our approach as NMC, for *Neural Model Checking* (it can also stand for "New Model Checking", as in a new approach to MC). Because of the possibility of false positives (FPs) and false negatives (FNs), NMC is best viewed as a solution to the *approximate model checking* (AMC) problem. Unlike previous work on AMC, however, we do not assume that the model is stochastic; see e.g. [39, 71]. Note that FPs and FNs are called Type I and Type II errors, respectively, in the theory of statistical hypothesis testing.

A well-trained NMC model checker offers a robust solution to the AMC problem, a solution that runs in constant time (approximately 1 millisecond, in our experiments) and takes constant space (an NN with one to three hidden layers and a reasonable number of neurons uses very little space). There are at least two use-cases for NMC: perform AMC on previously unseen states (i.e., states not in the dataset used for training); and for *online model checking*, where in the process of monitoring a system's behavior, one would like to determine, in real-time, the fate of the system going forward from the current state.

A common variant of the bounded-reachability problem considered above is where one is given a starting region $I$ instead of just a single starting state $s$. NMC can be extended to this case by applying output range estimation techniques that allow to compute estimated [46] or rigorous [28] bounds for the output of the NN on a given region of the input space.

Our NMC method comprises a pipeline of techniques that, in addition to the estimation of prediction accuracy, enable:

(1) The derivation of statistical guarantees to certify that the AMC meets prescribed levels of accuracy, FP and FN rates. This method, inspired by statistical model checking [71] and based on hypothesis testing, provides a simple, yet effective way to certify the performance of the AMC on unseen data, as opposed to neural network verification methods [31, 41, 45] that focus on the formal analysis of the network's output.

(2) Region-specific performance evaluation to assess how reliable is the AMC in specific sub-regions of the state-space, which is a crucial analysis for online model checking to identify in which states the AMC can be safely queried.

(3) Tuning of the learned AMC through adaptation (i.e., retraining with additional samples) or selection of the classification threshold. We will employ tuning to reduce the rate of false negatives, thus making the AMC more conservative.

Our experimental results demonstrate the feasibility and promise of this approach. In particular, we consider three well-established benchmarks in the hybrid systems community: a 2-variable spiking neuron, an inverted pendulum , and a 7-variable quadcopter controller. We consider shallow (1 hidden layer) and deep (3 hidden layers) NNs with sigmoid and ReLU activation functions, as well as two different NN ensembles. Applying these techniques on training and test datasets ranging in size from 5,000 to 20,000 samples, we achieve a prediction accuracy of 99.82% to 100% and an FN rate of 0.0007 to 0, taking into account the best-performing technique for each of the three benchmarks. We believe that such a range for the FN rate is acceptable in many practical applications and we show how this can be further improved through tuning of the classifiers.

In particular, we found that the deep NN classifiers yield superior accuracy compared to shallow NNs and other ML techniques, namely, support vector machines (SVMs) and binary decision trees (BDTs).

The rest of this paper develops along the following lines. Section 2 formally defines the AMC problem we are considering. Section 3 presents our NMC method. Section 4 describes the case studies used in our experimental evaluation. Section 4 presents our experimental results. Section 7 offers concluding remarks and directions for future work.

## 2 PROBLEM FORMULATION

We consider a general class of hybrid system models with continuous state-spaces and deterministic dynamics, possibly involving nonlinearities and jumps. Let $n$ be the number of state variables, $S \subseteq \mathbb{R}^n$ be the state space, and $\mathbb{T} \subseteq \mathbb{Q}^{\geq 0}$ be the time domain. A model $\mathcal{M}$ is a function $\mathcal{M} : S \times \mathbb{T} \to S$, such that, for state $s \in S$ and time $t \in \mathbb{T}$, $\mathcal{M}(s,t)$ is the state of the model after time $t$ starting from $s$. Let $S(\mathcal{M})$ denote the state space of a model $\mathcal{M}$.

We now formalize the problem of learning an approximate model checker from a set of examples (samples). We focus on time-bounded reachability properties, which check whether any state in a given set of states is reachable within some time horizon. Time-bounded reachability is well-suited for online model checking, which provides run-time safety guarantees for a fixed, relatively short time horizon.

*Definition 2.1 (Time-bounded reachability).* Given a model $\mathcal{M}$, a set of states $U \subseteq S(\mathcal{M})$, a state $s \in S(\mathcal{M})$, and a time bound $T \in \mathbb{T}$, decide whether there exists $t \leq T$ such that $\mathcal{M}(s,t) \in U$.

We consider a slightly relaxed notion of reachability, called *simulation-equivalent reachability* [5]. Intuitively, this captures reachability according to the discrete-time traces of the model, generated, for instance, using an ODE solver. For clarity, we assume fixed-step traces (i.e., all steps have the same duration), even though the definitions can easily be generalized to allow variable-step traces.

*Definition 2.2 (Simulation Trace).* Given a time step $h \in \mathbb{Q}^+$, the *simulation trace* of a model $\mathcal{M}$ from state $s \in S(\mathcal{M})$ and for time bound $T \in \mathbb{T}$, is the sequence of states

$$\rho_{\mathcal{M}}(s,T,h) = (\mathcal{M}(s,0), \mathcal{M}(s,h), \mathcal{M}(s,2h), \ldots, \mathcal{M}(s,kh)),$$

where $k = \lfloor T/h \rfloor$. We denote the length (number of states) of the trace with $|\rho_{\mathcal{M}}(s,T,h)|$. For $i \leq |\rho_{\mathcal{M}}(s,T,h)|$, we denote its $i$-th element with $\rho_{\mathcal{M}}(s,T,h)[i]$.

*Definition 2.3 (Simulation-equivalent time-bounded reachability).* Given a model $\mathcal{M}$, a set of states $U \subseteq S(\mathcal{M})$, a state $s \in S(\mathcal{M})$, a time bound $T \in \mathbb{T}$, and a time step $h \in \mathbb{Q}^+$, decide whether there exists $i \leq |\rho_{\mathcal{M}}(s,T,h)|$ such that $\rho_{\mathcal{M}}(s,T,h)[i] \in U$, denoted $\mathcal{M} \models \text{Reach}(U,s,T)$.

The time step $h$ is an implicit parameter of Reach. For brevity, we hereafter refer to simulation-equivalent time-bounded reachability simply as "reachability".

Note that our formulation allows arbitrarily complex system dynamics, provided the dynamics is deterministic. The dynamics itself can be a blackbox. We only require that there exists a procedure to decide $\mathcal{M} \models \text{Reach}(U,s,T)$ for a given model $\mathcal{M}$, state $s$, set of states $U$ and time bound $T$. We do not impose any specific language for expressing $\mathcal{M}$.

Before defining the problem of learning our approximate model checker, we describe the type of data from which it is learned. Let $\mathbb{B}$ denote the set of Boolean values.

*Definition 2.4 (Set of samples).* For model $\mathcal{M}$, set of states $U \subseteq S(\mathcal{M})$ and time bound $T \in \mathbb{T}$, a *set of samples* is any finite set:

$$\{(s,b) \in S(\mathcal{M}) \times \mathbb{B} \mid b = (\mathcal{M} \models \text{Reach}(U,s,T))\} \quad (1)$$

Thus, each sample consists of a state $s$ and a boolean $b$ which is the answer to the reachability problem starting from state $s$. We call $(s,1)$ a *positive sample* and $s$ a *positive state*. We call $(s,0)$ a *negative sample* and $s$ a *negative state*. Sets of samples are used for training and testing of the model checker.

Since each sample is labeled with the correct answer to the reachability problem instance, we have a supervised learning problem, specifically, a binary classification problem due to the Boolean categories.

Given a set of samples $D$, called the *training dataset*, the NMC *learning problem* is to learn a *classifier*, i.e., a total function $f : S \to \mathbb{B}$ from the training dataset. Learning typically corresponds to finding the parameters of the classifier function (weights and biases in the case of neural networks, see Section 3.1) that minimize some error function describing the discrepancy between training data and corresponding function predictions.

We do not require that the learned function agree with the training dataset $D$ on every state that appears in $D$. Imposing such a requirement can lead to over-fitting to $D$ and hence poor generalization to other states, lowering overall accuracy. We validate the learned function by assessing its behavior on a new dataset $D'$, called the *test dataset*, which is independent from the training dataset $D$. This is common practice in statistical analysis, especially when enough data is available to produce sufficiently large and independent training and test datasets. Other validation techniques, such as cross-validation [48], could also be employed.

We wish to evaluate the accuracy of the classifier $f$ in predicting the reachability values for the testing dataset $D'$. We consider three measures: overall *accuracy*, the rate of *false positives*, i.e., cases where $f$ incorrectly predicts that $U$ is reachable, and the rate of *false negatives*, i.e., cases where $f$ incorrectly predicts that $U$ is not reachable. Formally,

$$\text{Accuracy:} \quad \hat{P}_A = \frac{1}{n} \sum_{(s,b) \in D'} I(f(s) = b) \qquad (2)$$

$$\text{False positive rate:} \quad \hat{P}_{FP} = \frac{1}{n} \sum_{(s,b) \in D'} I(f(s) \wedge \neg b) \qquad (3)$$

$$\text{False negative rate:} \quad \hat{P}_{FN} = \frac{1}{n} \sum_{(s,b) \in D'} I(\neg f(s) \wedge b) \qquad (4)$$

where $n = |D'|$ and $I$ is the indicator function, which returns 1 if its argument is true, and 0 if its argument is false. In safety-critical applications where $U$ is a set of unsafe states, achieving a low false-negative rate is typically more important than achieving a low false-positive rate.

Note that accuracy, false positives and false negatives for a given test set $D'$ follow a Bernoulli distribution $B(1, p_x)$, where, for $x = $ A, FP, FN, $P_x$ denotes the true probability of success, which is estimated by the sample mean $\hat{P}_x$ (see Equations 2-4). The standard deviation is estimated by $\hat{\sigma}_x = \sqrt{\frac{\hat{P}_x \cdot (1 - \hat{P}_x)}{n}}$. For confidence level $\alpha > 0$, we can obtain the confidence interval $CI_x$ such that the real value of $P_x$ lies within $CI_x$ with probability $1 - \alpha$. We compute $CI_x$ using a Wilson-type interval [69], which is more reliable for extreme probabilities than the classical Wald-type intervals based on normal approximation. Indeed, accuracy, FN rate and FP rate typically take extreme probability values (close to 1, 0 and 0, respectively) when the classifier has good performance. The intervals are computed as follows:

$$CI_x = \frac{\hat{P}_x + \frac{z^2}{2n} \pm z \sqrt{\frac{\hat{P}_x \cdot (1 - \hat{P}_x)}{n} + \left(\frac{z}{2n}\right)^2}}{1 + \frac{z^2}{n}}, \qquad (5)$$

where $z = \Phi^{-1}(1 - \alpha/2)$ is the $(1 - \alpha/2)$-quantile of the standard normal distribution $\mathcal{N}(0, 1)$ (i.e., with mean 0 and standard deviation 1), where $\Phi$ is the cumulative distribution function of $\mathcal{N}(0, 1)$. In other words, $z$ tells the number of standard deviations away from the mean such that we cover $1 - \alpha/2$ of the probability of $\mathcal{N}(0, 1)$.

## 3 NEURAL MODEL CHECKING

Figure 1 illustrates a high-level schema of the Neural Model Checking method. As explained in Section 2 we start from a hybrid system model, which can be simulated to generate samples and populate training and testing datasets. Training data is used to learn the classifier, while test data to evaluate it. In Section 3.1, we provide background on (deep) neural network classifiers, while the sampling method is explained in Section 3.2. We stress that our method does not impose restrictions on the kind of classifier, and as we will see in the results section, we also support other machine learning models such as support vector machines and binary decision trees. For instance, instead of just one classifier, we can learn an ensemble of classifiers, that is, a classifier producing predictions based on e.g. majority voting or averaging of the predictions of multiple, possibly heterogeneous, classifiers. In our evaluation (see Section 5), we will consider two different ensembles of deep neural networks.

The learned classifier for model checking can be then analyzed to estimate its performance in terms of accuracy, false positive and false negative rates, which are estimated (together with their confidence intervals) from the test data (see Section 2). In addition to estimation, we can provide statistical guarantees using hypothesis testing (*a la* statistical model checking [71]) to certify that the classifiers meet prescribed performance levels (see Section 3.3). Region-specific analysis (Section 3.4) consists in evaluating the performance measures at a finer scale, i.e., locally to each state, thus providing a detailed picture of which state space sub-regions can be accurately predicted.

Finally, we consider two well-established methods to tune the classifier and improve its performance, illustrated in Section 3.5: adaptation, through which the classifiers are re-trained by incorporating wrongly predicted samples, in this sense being similar to well-established counterexample-guided approaches to model checking [16]; and threshold selection, i.e., adjusting the classification threshold to tune the error to favor either FNs or FPs. To make the classifier more conservative, we are more interested in reducing the rate of FNs, even though we can equally support other performance requirements.

### 3.1 Neural Networks for Classification

We use *feedforward* neural networks, a type of neural network that has one-way connections from input to output layers. Neural networks typically consist of several layers of neurons. We use *shallow* NNs which have one hidden layer connected to one output layer, and *deep* NNs which have more than one hidden layers. The neural networks are also fully connected, i.e., each neuron in a layer is *fully connected* to all neurons in the previous layer, as shown in Figure 2.

Let $l$ be the number of layers of the NN, i.e., $l - 1$ hidden and one output layers and let $n_i$ be the number of neurons in layer $i$, $i = 1, \ldots, l$, with $n_0$ being the size of the input vector.

For an input vector $x \in \mathbb{R}^{n_0}$, the output of the NN classifier is positive if $F(x) \geq \theta$, negative otherwise, where $F(x)$ is the function represented by the NN and $\theta$ is the classification threshold (see Section 3.5). Function $F$ is of the following form:

$$F = f_l \circ f_{l-1} \circ \ldots \circ f_1 \circ f_0,$$

where $\circ$ is the function composition operator, $f_0$ is the input normalization function, and for $i = 1, \ldots, l$, $f_i$ is the function computed by the $i$-th layer. The input normalization function typically applies a linear scaling such that the input falls in the range $[-1, 1]$:

$$f_0(x) = -1 + 2 \cdot (x - x_{\min}) \oslash (x_{\max} - x_{\min}) \qquad (6)$$

where $\oslash$ is the Hadamard (a.k.a. entrywise) division, $x_{\min}$ and $x_{\max}$ are respectively the vectors of minimum and maximum components over all the training dataset.

The output of layer $i$ results from the application of function $f_i : \mathbb{R}^{n_{i-1}} \to \mathbb{R}^{n_i}$ to the output of the previous layer:

$$f_i(p_{i-1}) = g_i(W_{i,i-1} \cdot p_{i-1} + b_i), \quad i = 1..l \qquad (7)$$

where $p_{i-1} \in \mathbb{R}^{n_{i-1}}$ is the output vector of layer $i - 1$, $W_{i,i-1} \in \mathbb{R}^{n_i \times n_{i-1}}$ is the *weight matrix* that connects $p_{i-1}$ to the neurons of layer $i$, $b_i \in \mathbb{R}^{n_i}$ is the *bias vector* of layer $i$, and $g_i$ is the *activation function* of the neurons of layer $i$.

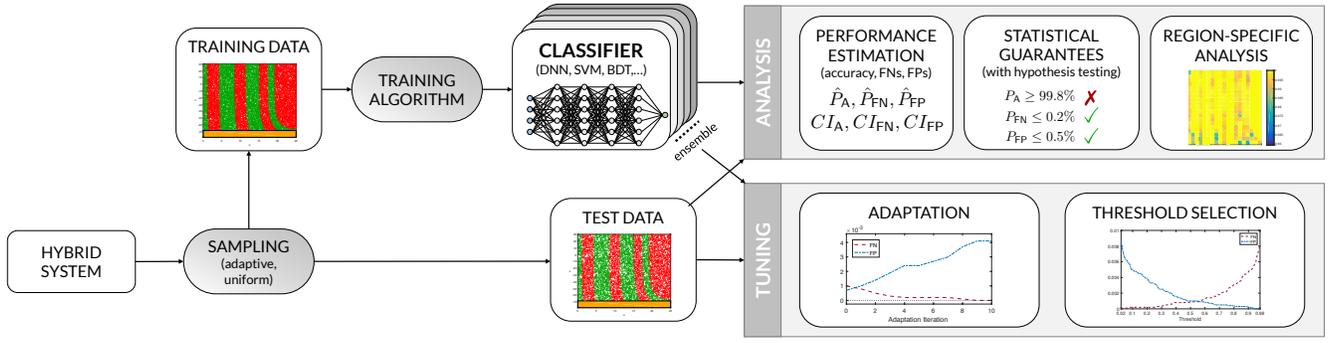

Figure 1: Diagram of the Neural Model Checking method

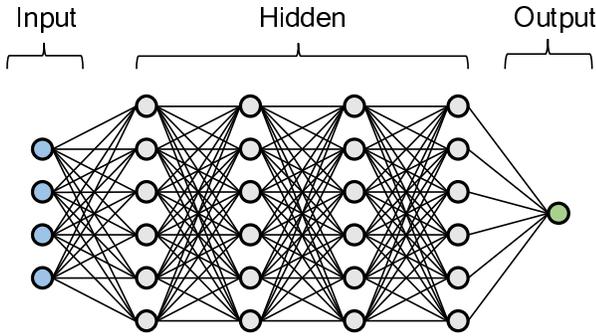

Figure 2: A fully connected feedforward neural network with 4 inputs, 4 hidden layers, and 1 output.

Weights and biases are the function parameters learned during training, and are typically derived by minimizing the mean square error (or other error functions) between training data and network predictions. The most common optimization algorithm is gradient descent with backpropagation [21, 38].

In our evaluation, we will consider two main configurations of NNs (see also Section 5): the first, called DNN-S, uses the Tan-Sigmoid activation function tansig for the hidden layers and the Log-Sigmoid activation function logsig for the output layer $l$. Let $z \in \mathbb{R}^{n_i}$ be the argument of the activation function at layer $i$. Then, for neuron $j = 1, \ldots, n_i$, the above activation functions are given by:

$$\text{tansig}(z)_j = \frac{2}{1 + e^{-2 \cdot z_j}} - 1 \quad \text{and} \quad \text{logsig}(z)_j = \frac{1}{1 + e^{-z_j}}. \quad (8)$$

The second configuration, called DNN-R, employs the rectified linear unit (ReLU) activation function relu for the hidden layers and the softmax function for the output layer $l$, where

$$\text{relu}(z)_j = \max(0, z_j) \quad \text{and} \quad \text{softmax}(z)_j = \frac{e^{z_j}}{\sum_{k=1}^{n_i} e^{z_k}}. \quad (9)$$

### 3.2 Generation of Training Data and Test Data

Given a model $\mathcal{M}$ with state-space $S(\mathcal{M})$, a set of states $U \subset S(\mathcal{M})$, a time bound $T \in \mathbb{T}$, and time step $h \in \mathbb{Q}^+$, we generate data for training and testing as follows. We select a state $s \in S(\mathcal{M})$ by sampling from an appropriate distribution (as discussed below) and simulate $\mathcal{M}$ starting from $s$ using time step $h$ until the time bound $T$ is reached, to obtain a simulation trace $\rho_{\mathcal{M}}(s, T, h)$. We then classify $s$ as either positive or negative, depending on whether $\rho_{\mathcal{M}}(s, T, h)$ contains a state in $U$, as per Definition 2.3. We repeat this process until the specified number of samples is generated. For test data, we use a uniform distribution to sample $s$ from $S(\mathcal{M})$, to obtain an unbiased evaluation.

For training data, we observed that in applications where the unsafe states $U$ are a small part of the overall state space, a uniform sampling strategy produces unbalanced training datasets that contain insufficient positive samples, causing the learned classifier to have relatively low accuracy. We address this problem by using an *adaptive* sampling strategy. In this strategy, we uniformly sample states $s$ from $S(\mathcal{M})$, but when we get a positive sample, we generate an additional $n$ samples by sampling in a small region around $s$. The value of $n$ is application-specific and is chosen such that the generated dataset contains comparable numbers of positive and negative samples.

### 3.3 A Posteriori Statistical Guarantees

It is well-known that training deep neural networks with guaranteed performance is still an unsolved problem. For this reason, we propose to provide performance guarantees a posteriori, i.e., after training. Inspired by statistical approaches to model checking [71], we employ statistical hypothesis testing to certify our classifiers for model checking by providing statistical guarantees on accuracy, false positive and false negative rates. Corresponding results are reported in Section 5.2.

In particular, we provide guarantees of the form $P_A \geq \theta_A$ (i.e., the true accuracy value is above $\theta_A$), $P_{FN} \leq \theta_{FN}$ and $P_{FP} \leq \theta_{FP}$ (i.e., the true rate of FNs and FPs are respectively below $\theta_{FN}$ and $\theta_{FP}$). Being based on hypothesis testing, such guarantees are precise up to arbitrary error bounds $\alpha, \beta \in (0, 1)$, such that the probability of Type-I errors (i.e., for $x = A, FN, FP$, of accepting $P_x < \theta_x$ when $P_x \geq \theta_x$) is bounded by $\alpha$, and the probability of of Type-II errors (i.e., for $x = A, FN, FP$, of accepting $P_x \geq \theta_x$ when $P_x < \theta_x$) is bounded by $\beta$. The pair $(\alpha, \beta)$ is known as the *strength* of the test.

To ensure both error bounds simultaneously, the original test $P_x \geq \theta_x$ vs $p_x < \theta_x$ is relaxed by introducing a small indifference region, i.e., we test the hypothesis $H_0 : P_x \geq p_0$ against $H_1 : P_x \leq$ $p_1$, with $p_0 > p_1$ [71]. Typically, $p_0 = \theta_x + \delta$ and $p_1 = \theta_x - \delta$ for some $\delta > 0$. We use Wald's sequential probability ratio test (SPRT) [66] to provide the above guarantees. SPRT has the important advantage that it does not require a prescribed number of samples to accept one of the two hypothesis, but the decision is made if the available samples provide sufficient evidence. Specifically, after $m$ samples, hypothesis $H_0$ is accepted if $\frac{p_{1m}}{p_{0m}} \leq B$, while hypothesis $H_1$ is accepted if $\frac{p_{1m}}{p_{0m}} \geq A$, where $A = (1-\beta)/\alpha$, $B = \beta/(1-\alpha)$ and $\frac{p_{1m}}{p_{0m}} = \frac{p_1^{t_m} \cdot (1-p_1)^{f_m}}{p_0^{t_m} \cdot (1-p_0)^{f_m}}$, where $t_m$ and $f_m$ are, respectively, the numbers of positive and negative samples in the current set of $m$ samples ($t_m = m - f_m$).

We remark that the computation of confidence intervals (explained in Section 2) provides *per se* a kind of statistical guarantee, but their purpose is to identify an interval containing the true probability value with high probability. In contrast, the above guarantees based on statistical hypothesis testing focus on certifying that the classifiers meet given performance levels, as in statistical model checking.

### 3.4 Region-specific analysis
Motivated by online model checking applications, where predictions about reachability of a bad state are made at runtime from the current state, it is important to evaluate the performance of the classifiers at a finer scale, i.e., locally to each state.

In other words, we perform statistical analysis to estimate accuracy, false negatives and false positives by generating test datasets from small sub-regions of the state space. Such an analysis gives a detailed view of the regions with better prediction accuracy and allows spotting "problematic" regions with poor prediction performance, thus prompting countermeasures focused to the problematic state space regions, such as additional training or adaptation, tuning of the classification threshold (see Section 3.5) or replacing the classifier with a certified reachability checker.

In Section 5.3, we provide a detailed region-specific performance evaluation for our case studies, showing that for the largest part of the state space, our neural network classifier yields very precise results with 100% accuracy, with acceptable accuracy even for the "problematic" regions.

### 3.5 Reducing False Negatives through Threshold Selection and Adaptation
As explained in Section 3.1, the NN classifier is based on a classification threshold $\theta$, as are other kinds of classifiers. This threshold is typically set to 0.5, such that network predictions are classified as either negative or positive depending on whether or not the prediction is below the threshold. However, in many situations where, for instance, the testing data is imbalanced, the natural choice of $\theta = 0.5$ is not suitable, and improved accuracy can be achieved through the analysis of different classification thresholds [73].

There is an inevitable tradeoff between FN and FP rates: by decreasing $\theta$, we reduce the number of false negatives because the classifier will tend to answer in a positive way to additional inputs, but for the same reason, we increase the number of false positives. Keeping in mind that false negatives are the most serious errors from a safety-critical perspective, a threshold selection strategy that increases FPs to a larger extent than it reduces FNs might still be viable, even though extreme thresholds (close to 0 or 1) typically lead to catastrophic loss of accuracy. In Section 5.5, we show different threshold selection strategies able to considerably reduce the FN rate.

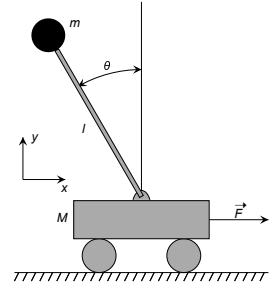

**Figure 3: Schematic of the inverted pendulum on a cart. Source: Wikipedia.**

Another way to reduce false negatives of a NN classifier is *adaptation*, in which a set of additional samples is used to update the weights and/or biases of a previously trained neural network, in this way enabling incremental retraining of the classifier. This technique shares similarities with the well-established model-checking method of counterexample-guided abstraction refinement (CEGAR) [16] in that we also use counterexamples to adapt the classifier. Unlike CEGAR where spurious counterexamples trigger a refinement step that makes the model less conservative, in our adaptation, retraining with false negatives makes the classifier more conservative, as we will show in Section 5.4.

## 4 MODELS AND CASE STUDIES
### 4.1 Inverted Pendulum
We consider the control system for an inverted pendulum on a cart. This is a classic, widely used example of a non-linear system. As shown in Fig. 3, the control input $F$ is a force applied to the cart with the goal of keeping the pendulum in upright position, i.e., $\theta = 0$. The dynamics is given by

$$J \cdot \ddot{\theta} = m \cdot l \cdot g \cdot \sin(\theta) - m \cdot l \cos(\theta) \cdot F \tag{10}$$

Following [15], we set $J = 1$, $m = 1/g$, $l = 1$, and let $u = F/g$. Eq. 10 becomes

$$\begin{cases} \dot{\theta} = \omega \\ \dot{\omega} = \sin(\theta) - \cos(\theta) \cdot u \end{cases} \tag{11}$$

We consider the control law given in [15] and shown in Eq. 12. Fig. 4 shows an evolution of $\theta$ under this control law. We consider the unsafe state set $U = \{(\theta, \omega) \mid \theta < -\pi/4 \lor \theta > \pi/4\}$. This unsafe region corresponds to the safety property that keeps the pendulum within 45° of the vertical axis.

Datasets for training and test and reported in Table 1 and illustrated in Figure 5. The domain for sampling is $\theta \in [-\pi/4, \pi/4] \land \omega \in [-1.5, 1.5]$. We used time bound $T = 5$.

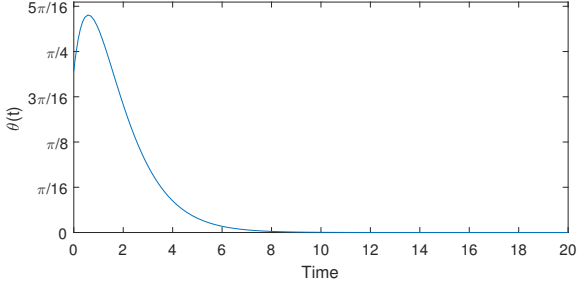

Figure 4: An evolution of the inverted pendulum state variable $\theta$ from initial state $(\theta_0, \omega_0) = (0.5, 1.0)$.

| Dataset ID | # samples | % positive | Strategy | Use |
|---|---|---|---|---|
| IP-DS-1 | 10,000 | 34.85% | Adaptive | Training |
| IP-DS-2 | 20,000 | 40.8% | Adaptive | Training |
| IP-DS-3 | 10,000 | 12.5% | Uniform | Test |

Table 1: Training and test datasets for the inverted pendulum model. *% positive*: proportion of positive samples. *Strategy*: sampling strategy. *Use*: training or test.

$$u = \begin{cases} \dfrac{2 \cdot \omega + \theta + \sin(\theta)}{\cos(\theta)}, & E \in [-1, 1], |\omega| + |\theta| \le 1.85 \\ 0, & E \in [-1, 1], |\omega| + |\theta| > 1.85 \\ \dfrac{\omega}{1+|\omega|}\cos(\theta), & E < -1 \\ \dfrac{-\omega}{1+|\omega|}\cos(\theta), & E > 1 \end{cases} \quad (12)$$

where $E = 0.5 \cdot \omega + (\cos(\theta) - 1)$ is the pendulum energy.

## 4.2 Spiking Neuron

We consider the spiking neuron model on the Flow* website[1], which is based on a model in [43]. It is a hybrid system with one mode and one jump. The dynamics is defined by the ODE

$$\begin{cases} \dot{v} = 0.04v^2 + 5v + 140 - u + I \\ \dot{u} = a \cdot (b \cdot v - u) \end{cases} \quad (13)$$

The jump condition is $v \ge 30$, and the associated reset is $v' := c \wedge u' := u + d$, where, for any variable $x$, $x'$ denotes the value of $x$ after the reset.

The parameters are $a = 0.02, b = 0.2, c = -65, d = 8$, and $I = 40$ as reported on the Flow* website. We consider the unsafe state set $U = \{(v, u) \mid v \le 68.5\}$. This corresponds to a safety property that can be understood as the neuron does not undershoot its resting-potential region of $[-68.5, -60]$. Fig. 6 shows an example evolution of $v$.

Datasets for training and test and reported in Table 2 and illustrated in Figure 7. The domain for sampling is $68.5 < v \le 30 \wedge 0 \le$

[1]https://flowstar.org/examples/

| Dataset ID | # samples | % positive | Strategy | Use |
|---|---|---|---|---|
| SN-DS-1 | 10,000 | 53.97% | Uniform | Training |
| SN-DS-2 | 20,000 | 54.02% | Uniform | Training |
| SN-DS-3 | 10,000 | 54.73% | Uniform | Test |

Table 2: Training and test datasets for the spiking neuron model. For this model, uniform sampling yields a good balance between positive and negative samples and thus adaptive sampling was not required.

$u \le 25$. The time bound for the reachability property was set to $T = 20$.

## 4.3 Quadcopter Controller

We consider the quadcopter model used as a benchmark for dReal [1]. We consider the safety property that the quadcopter does not crash, i.e., the altitude $z$ is positive. This corresponds to the unsafe state set $U$ defined by $z \le 0$. This safety property is independent of the state variables $x$, $y$, and $\psi$ (the yaw angle), so we omit them from the model. This hybrid system has two modes that share the following ODEs.

$$\begin{cases} \dfrac{d\omega_x}{dt} = \dfrac{L \cdot k \cdot (\omega_1^2 - \omega_3^2) - (I_{yy} - I_{zz}) \cdot \omega_y \cdot \omega_z}{I_{xx}} \\ \dfrac{d\omega_y}{dt} = \dfrac{L \cdot k \cdot (\omega_2^2 - \omega_4^2) - (I_{zz} - I_{xx}) \cdot \omega_x \cdot \omega_z}{I_{yy}} \\ \dfrac{d\omega_z}{dt} = \dfrac{b \cdot (\omega_1^2 - \omega_2^2 + \omega_3^2 - \omega_4^2) - (I_{xx} - I_{yy}) \cdot \omega_x \cdot \omega_y}{I_{zz}} \\ \dfrac{d\phi}{dt} = \omega_x + \dfrac{\sin(\phi)\sin(\theta)}{\left(\dfrac{\sin(\phi)^2\cos(\theta)}{\cos(\phi)} + \cos(\phi)\cos(\theta)\right)\cos(\phi)}\omega_y \\ \qquad + \dfrac{\sin(\theta)}{\dfrac{\sin(\phi)^2\cos(\theta)}{\cos(\phi)} + \cos(\phi)\cos(\theta)}\omega_z \\ \dfrac{d\theta}{dt} = -\left(\dfrac{\sin(\phi)^2\cos(\theta)}{\left(\dfrac{\sin(\phi)^2\cos(\theta)}{\cos(\phi)}\omega_y + \cos(\phi)\cos(\theta)\right)\cos(\phi)^2} + \dfrac{1}{\cos(\phi)}\right)\omega_y \\ \qquad - \dfrac{\sin(\phi)\cos(\theta)}{\left(\dfrac{\sin(\phi)^2\cos(\theta)}{\cos(\phi)} + \cos(\phi)\cos(\theta)\right)\cos(\phi)}\omega_z \\ \dfrac{dz}{dt} = \dot{z} \end{cases} \quad (14)$$

where the dynamics of $z$ is given by:

$$(\text{mode 1}) \quad \dfrac{d\dot{z}}{dt} = \dfrac{g + \cos(\theta) \cdot k \cdot (\omega_1^2 + \omega_2^2 + \omega_3^2 + \omega_4^2) + k \cdot d \cdot \dot{z}}{m} \quad (15)$$

$$(\text{mode 2}) \quad \dfrac{d\dot{z}}{dt} = \dfrac{-g - \cos(\theta) \cdot k \cdot (\omega_1^2 + \omega_2^2 + \omega_3^2 + \omega_4^2) - k \cdot d \cdot \dot{z}}{m} \quad (16)$$

The jump from mode 1 to mode 2 happens when $z = 500$, updating variables to $\omega_1' := 0 \wedge \omega_2' := 1 \wedge \omega_3' := 0 \wedge \omega_4' := 1$. The jump from mode 2 to mode 1 occurs at $z = 200$, updating variables to $\omega_1' := 1 \wedge \omega_2' := 0 \wedge \omega_3' := 1 \wedge \omega_4' := 0$.

Following [1], the parameters are $L = 0.23, k = 5.2, k \cdot d = 7.5\text{e}{-7}$, $m = 0.65, b = 3.13\text{e}{-5}, g = 9.8, I_{xx} = 0.0075, I_{yy} = 0.0075$, $I_{zz} = 0.013$. Fig. 8 shows an example evolution of $z$. Datasets for training and test and reported in Table 1. The domain for sampling is $\omega_x \in [-0.05, 0.05], \omega_y \in [0, 0.1], \omega_z \in [-0.1, 0.1], \phi \in [-0.2, 0.2]$,

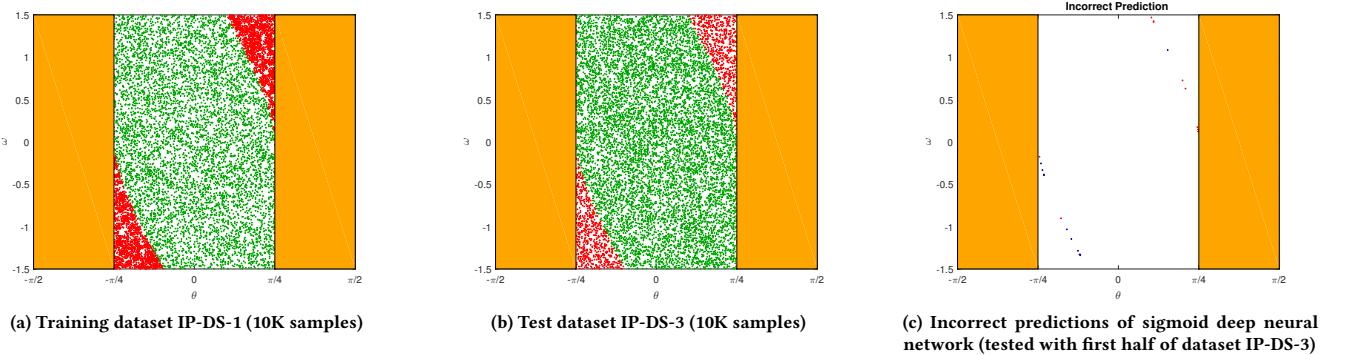

(a) Training dataset IP-DS-1 (10K samples)  (b) Test dataset IP-DS-3 (10K samples)  (c) Incorrect predictions of sigmoid deep neural network (tested with first half of dataset IP-DS-3) $\theta \in [-1, 0.4]$, $\dot{z} \in [-150, 150]$, and $z \in [50, 100]$. We chose time bound $T = 15$.

Figure 5: Training dataset (a), test dataset (b) and incorrect predictions (c) for the inverted pendulum model. The orange area is the unsafe region. In plots (a,b), green dots are negative samples and red dots are positive samples. In plot (c), blue dots are false positives and red dots are false negatives.

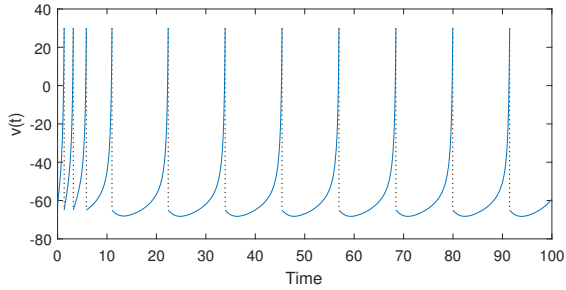

Figure 6: An evolution of the spiking-neuron state variable $v$ from initial state $(v_0, u_0) = (-62, 0.1)$. The dotted lines represent discontinuities caused by jumps.

| Dataset ID | # samples | % positive | Strategy | Use |
|---|---|---|---|---|
| QC-DS-1 | 10,000 | 47.47% | Adaptive | Training |
| QC-DS-2 | 20,000 | 46.99% | Adaptive | Training |
| QC-DS-3 | 10,000 | 72.19% | Uniform | Test |

Table 3: Training and test datasets for the spiking neuron model.

## 5 RESULTS

In this section, we evaluate the performance (accuracy, FNs and FPs) of the classifiers for model checking for the three case studies (Section 5.1). We further illustrate the analysis and tuning methods at the core of our Neural Model Checking method (see Section 3). Namely, we provide statistical guarantees on the derived classifiers (Section 5.2); evaluate local performance by examining smaller state space regions (Section 5.3); and show how to drastically reduce the FN rate by means of adaptation (Section 5.4) and threshold selection (Section 5.5). Finally, we also analyze the impact of different time bounds in the reachability property (Section 5.6).

For all case studies, neural networks are learned with MATLAB's train function. Specifically we employ the Levenberg-Marquardt [21, 38] backpropagation training algorithm with the mean square error performance function, and the Nguyen-Widrow [54] initialization method for the NN layers. Training is very fast, taking 1 to 9 seconds for a training dataset with 10,000 samples and 2 to 19 seconds for a training dataset with 20,000 samples.

In our evaluation we compare deep and shallow neural networks with alternative classifiers. In particular, for each training dataset, we learned the following classifiers:

- A sigmoid deep neural network (**DNN-S**) with 3 hidden layers of 10 neurons each and one output layer. The hidden layers use tansig, and the output layer uses logsig as activation functions.
- A ReLU deep neural network (**DNN-R**) with 3 hidden layers of 10 neurons each and one output layer. The hidden layers use relu, and the output layer uses softmax as activation functions.
- A shallow neural network (**SNN**) with one hidden layer of 20 neurons and one output layer. The hidden layer uses tansig, and the output layer uses logsig as activation functions.
- A support vector machine (**SVM**) with a radial kernel.
- A binary decision tree (**BDT**).
- An ensemble of five sigmoid DNNs (**Ens1**) trained with different datasets. The result of the classification is given by majority voting.
- An ensemble of three sigmoid DNNs and two ReLU DNNs (**Ens2**).

To evaluate the effect of different sizes for the training set, for each of the above classifiers we trained two variants: 1) using the 10K-sample datasets for training and half of the 10K-sample test datasets; 2) using the 20K-sample datasets for training and the full 10K-sample test datasets. Training data for the network ensembles were generated with consistent sampling strategies and number of samples.

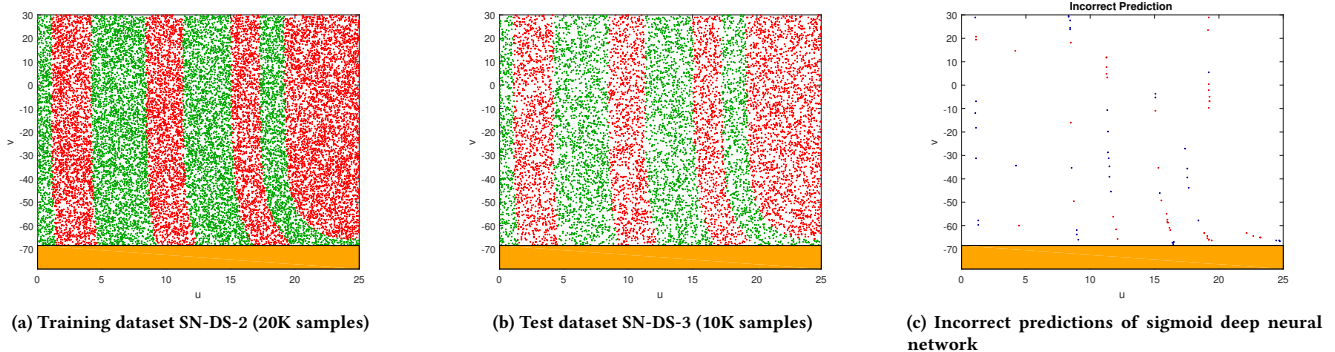

(a) Training dataset SN-DS-2 (20K samples)  (b) Test dataset SN-DS-3 (10K samples)  (c) Incorrect predictions of sigmoid deep neural network

Figure 7: Training dataset (a), test dataset (b) and incorrect predictions (c) for the spiking neuron model. Color code is the same as Figure 5

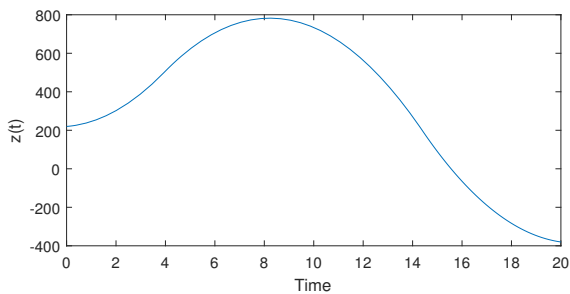

Figure 8: An evolution of $z$ leading to a quadcopter crash.

The number of layers and number of neurons are chosen empirically. To avoid overfitting, we did not try to choose a number that achieves the best result on the test dataset. All models were simulated using MATLAB's ode45 variable-step ODE solver.

### 5.1 Performance evaluation

Table 4 shows the performances of all classifiers for the three case studies. In all case studies, the ensemble of classifiers **Ens1** has the best accuracy and false negative rate, with the ensemble **Ens2** performing slightly better in terms of false positive rate. If we consider the individual classifiers, the sigmoid DNN **DNN-S** has the best overall performances among the classifiers trained with 10K samples, second only to the shallow neural network **SNN** for the FN rate of the quadcopter model. Among the individual classifiers trained with 20K samples, **DNN-S** yields the best results for the spiking neuron and inverted pendulum models while **DNN-R** is the best classifier for the quadcopter controller. In general we find that the NN-based classifiers has superior performance compared to support vector machines and binary decision trees.

Overall, the best classifiers for the three case studies achieve accuracy levels ranging from 99.82% to 100% and false negative rates of 0.07% to 0%. As the false negative rate can be further improved by adaptation and threshold selection (see results in Section 5.4 and Section 5.5), we believe that this level of accuracy is acceptable in many practical applications. Importantly, the classifiers yield very tight 99% confidence intervals, meaning that our estimation of accuracy, FN and FP is sufficiently precise.

As shown in Fig. 5 (c) and Fig. 7 (c), FN and FP samples are concentrated at the border between the positive and negative region, as confirmed also by the local analysis of Section 5.3. In Section 5.4, we show that adaptation can shift the decision boundary of the NN to reduce FNs at the cost of a slight increase in FPs.

### 5.2 A Posteriori Statistical Guarantees

We provide statistical guarantees using hypothesis testing (as explained in Section 5.2) for all models and classifiers (only the variants trained with 20K samples). Results are reported in Table 5 and obtained with $\alpha = \beta = 0.01$ and $\delta = 0.001$. We assess six properties, given by $P_A \geq 99.5\%$, 99.8%, $P_{FN} \leq 0.5\%$, 0.2% and $P_{FP} \leq 0.5\%$, 0.2%. We report that the only classifier able to satisfy all six properties is the ensemble of sigmoid DNNs. However, the single DNN and the mixed ensemble of DNNs have comparable performance and fail only for property $P_A \geq 99.8\%$ for the neuron model. In accordance with the results of Table 4, the neuron model is the hardest to predict for our classifiers, followed by the quadcopter and pendulum models.

Crucially, this analysis evidences that only a small number of samples are required to obtain statistical guarantees with the given strength, making it suitable to provide run-time assurance in online model checking scenarios. Indeed only 10 out of 126 tests needed more than 10K samples to reach a decision, with 11 tests terminated with less than 1K samples.

### 5.3 Region-specific analysis

To evaluate region-specific performance, we estimate accuracy, false negatives and false positives (and corresponding confidence intervals) in smaller sub-regions of the state space, as explained in Section 3.4. We performed this analysis for the pendulum and neuron case studies considering the DNN classifier (trained with 20K samples), see results in Figure 9. We divided the 2-dimensional state spaces in a 20×20 grid, generating a test dataset of 10,000 uniform samples for each grid cell.

### 10K sample training set, pendulum

|       | Acc                    | FN                   | FP                   |
|-------|------------------------|----------------------|----------------------|
| DNN-S | **100** [99.867,100]   | **0** [0.000,0.133]  | **0** [0.000,0.133]  |
| DNN-R | 99.92 [99.731,99.977]  | 0.02 [0.002,0.171]   | 0.06 [0.015,0.238]   |
| SNN   | 99.8 [99.558,99.91]    | 0.18 [0.078,0.414]   | 0.02 [0.002,0.171]   |
| SVM   | 99.74 [99.477,99.871]  | 0.24 [0.116,0.496]   | 0.02 [0.002,0.171]   |
| BDT   | 99.2 [98.804,99.466]   | 0.52 [0.315,0.856]   | 0.28 [0.142,0.55]    |
| Ens1  | **100** [99.867,100]   | **0** [0,0.133]      | **0** [0,0.133]      |
| Ens2  | 99.98 [99.829,99.998]  | 0.02 [0.002,0.171]   | **0** [0,0.133]      |

### 20K training samples, pendulum

|       | Acc                   | FN                   | FP                  |
|-------|-----------------------|----------------------|---------------------|
| DNN-S | 99.99 [99.914,99.999] | 0.01 [0.001,0.086]   | 0 [0,0.067]         |
| DNN-R | 99.9 [99.779,99.955]  | 0.07 [0.027,0.179]   | [0.007,0.119]       |
| SNN   | 99.77 [99.609,99.865] | 0.2 [0.113,0.353]    | 0.03 [0.007,0.119]  |
| SVM   | 99.83 [99.685,99.909] | 0.17 [0.091,0.315]   | **0** [0,0.067]     |
| BDT   | 99.6 [99.401,99.733]  | 0.23 [0.135,0.391]   | 0.17 [0.091,0.315]  |
| Ens1  | **100** [99.933,100]  | **0** [0,0.067]      | **0** [0,0.067]     |
| Ens2  | **100** [99.933,100]  | **0** [0,0.067]      | **0** [0,0.067]     |

### 10K sample training set, neuron

|       | Acc                   | FN                   | FP                   |
|-------|-----------------------|----------------------|----------------------|
| DNN-S | 99.6 [99.295,99.774]  | 0.22 [0.103,0.469]   | 0.18 [0.078,0.414]   |
| DNN-R | 99.06 [98.637,99.353] | 0.5 [0.3,0.831]      | 0.44 [0.255,0.756]   |
| SNN   | 98.48 [97.965,98.866] | 0.64 [0.407,1.003]   | 0.88 [0.598,1.292]   |
| SVM   | 98.04 [97.467,98.485] | 1.02 [0.713,1.457]   | 0.94 [0.647,1.363]   |
| BDT   | 98.32 [97.783,98.729] | 0.84 [0.566,1.244]   | 0.84 [0.566,1.244]   |
| Ens1  | 99.74 [99.477,99.871] | **0.1** [0.033,0.299]| 0.16 [0.066,0.386]   |
| Ens2  | 99.7 [99.424,99.844]  | 0.2 [0.09,0.442]     | **0.1** [0.033,0.299]|

### 20K training samples, neuron

|       | Acc                    | FN                   | FP                   |
|-------|------------------------|----------------------|----------------------|
| DNN-S | 99.81 [99.66,99.894]   | 0.09 [0.039,0.208]   | 0.1 [0.045,0.221]    |
| DNN-R | 99.52 [99.306,99.669]  | 0.18 [0.098,0.328]   | 0.29 [0.18,0.466]    |
| SNN   | 99.17 [98.901,99.374]  | 0.4 [0.267,0.599]    | 0.43 [0.291,0.635]   |
| SVM   | 98.73 [98.407,98.988]  | 0.52 [0.364,0.741]   | 0.75 [0.558,1.008]   |
| BDT   | 99.3 [99.05,99.485]    | 0.33 [0.211,0.515]   | 0.37 [0.243,0.563]   |
| Ens1  | **99.82** [99.672,99.902] | **0.07** [0.027,0.179] | 0.11 [0.051,0.235] |
| Ens2  | **99.82** [99.672,99.902] | 0.08 [0.033,0.194]   | **0.1** [0.045,0.221] |

### 10K training samples, quadcopter

|       | Acc                    | FN                   | FP                   |
|-------|------------------------|----------------------|----------------------|
| DNN-S | 99.8 [99.558,99.91]    | 0.06 [0.015,0.238]   | 0.18 [0.054,0.358]   |
| DNN-R | 99.7 [99.424,99.844]   | 0.1 [0.033,0.299]    | 0.2 [0.09,0.442]     |
| SNN   | 99.78 [99.531,99.897]  | **0.04** [0.007,0.205] | 0.24 [0.078,0.414] |
| SVM   | 97.04 [96.357,97.598]  | 2.34 [1.849,2.958]   | 0.62 [0.392,0.979]   |
| BDT   | 99.4 [99.045,99.624]   | 0.2 [0.09,0.442]     | 0.4 [0.226,0.705]    |
| Ens1  | **99.84** [99.614,99.934] | **0.04** [0.007,0.205] | **0.12** [0.043,0.329] |
| Ens2  | 99.64 [99.346,99.802]  | 0.16 [0.066,0.386]   | 0.2 [0.09,0.442]     |

### 20K training samples, quadcopter

|       | Acc                    | FN                   | FP                   |
|-------|------------------------|----------------------|----------------------|
| DNN-S | 99.83 [99.685,99.909]  | 0.07 [0.027,0.179]   | 0.1 [0.045,0.221]    |
| DNN-R | 99.89 [99.765,99.949]  | 0.05 [0.016,0.15]    | 0.06 [0.021,0.165]   |
| SNN   | 99.85 [99.711,99.922]  | 0.07 [0.027,0.179]   | 0.08 [0.033,0.194]   |
| SVM   | 97.33 [96.882,97.715]  | 1.98 [1.651,2.372]   | 0.69 [0.507,0.939]   |
| BDT   | 99.52 [99.306,99.669]  | 0.28 [0.172,0.453]   | 0.2 [0.113,0.353]    |
| Ens1  | **99.93** [99.821,99.973] | **0.01** [0.001,0.086] | 0.06 [0.021,0.165] |
| Ens2  | 99.91 [99.792,99.961]  | 0.04 [0.011,0.135]   | **0.05** [0.016,0.15] |

Table 4: Accuracy (Acc), FP rate, and FN rate of the learned classifier for each case study, classifier type, and training dataset size. All results are expressed as percentages and are reported as $a\,[b,c]$, where $a$ is the sample mean and $[b,c]$ is the 99% confidence interval (conservative over-approximation to the closest decimal). For each measure and each training dataset, the best result is highlighted in bold.

Such analysis confirms that the most problematic regions are found at the decision borders (compare with Figures 5 c and 7 c). Nevertheless, we observe that most of the regions yield 100% accuracy, with all 99% confidence intervals contained in $[0.9697, 1]$ for the pendulum model and in $[0.9592, 1]$ for the neuron model. Similarly, false negative and positive rates are largely equal to 0. The 99% confidence intervals for the FN and FP rates are all contained in $[0, 0.019]$ and $[0, 0.0303]$ respectively for the pendulum model, and in $[0, 0.0376]$ and $[0, 0.0408]$ for the neuron model.

### 5.4 Reducing False Negatives through Adaptation

In this section, we evaluate the benefits of adaptation by incrementally adapting the trained NNs with false negative samples (see Section 3.5). The adaptation experiments were performed for each case study on the sigmoid DNN trained with 20K samples as follows. At each iteration, we generate a different 10K-sample dataset, which we use to test the current network. The network is then adapted with the corresponding set of FN samples. Note that the performance of the adapted NN reported in Figure 10 is measured against the original 10K-sample test dataset.

We employ MATLAB's adapt function with gradient descent learning algorithm and a learning rate of 0.001, 0.0005, and 0.002 for the inverted pendulum, spiking neuron, and quadcopter controller, respectively. We remark that in our case studies, adapting only the layer weights produces the best results. Fig. 10 shows the adaptation results for all case studies. In the spiking neuron and quadcopter case studies, adaptation helps decrease the FN rates to 0% at the cost of a slight increase in the FP rates. In the inverted pendulum case study, the DNN already has a FN rate of 0% on the original test dataset (see also Table 4). It also has a FN rate of 0% on 6 of the 10 test datasets used for the incremental adaptation. As a result, adaptation is not effective for this case study, since it keeps the FN rate at 0% while increasing the FP rate.

Figure 11 visualizes the effects of adaptation on the DNN **DNN-S** originally trained with 20K samples for the spiking neuron case study. Fig. 11 (a) shows the prediction of the DNN after training with 20K samples. Fig. 11 (b) shows the prediction of the DNN after being adapted with a total of 31 negative samples spread over 9 iterations. It can be seen that after adaptation, the predicted positive region

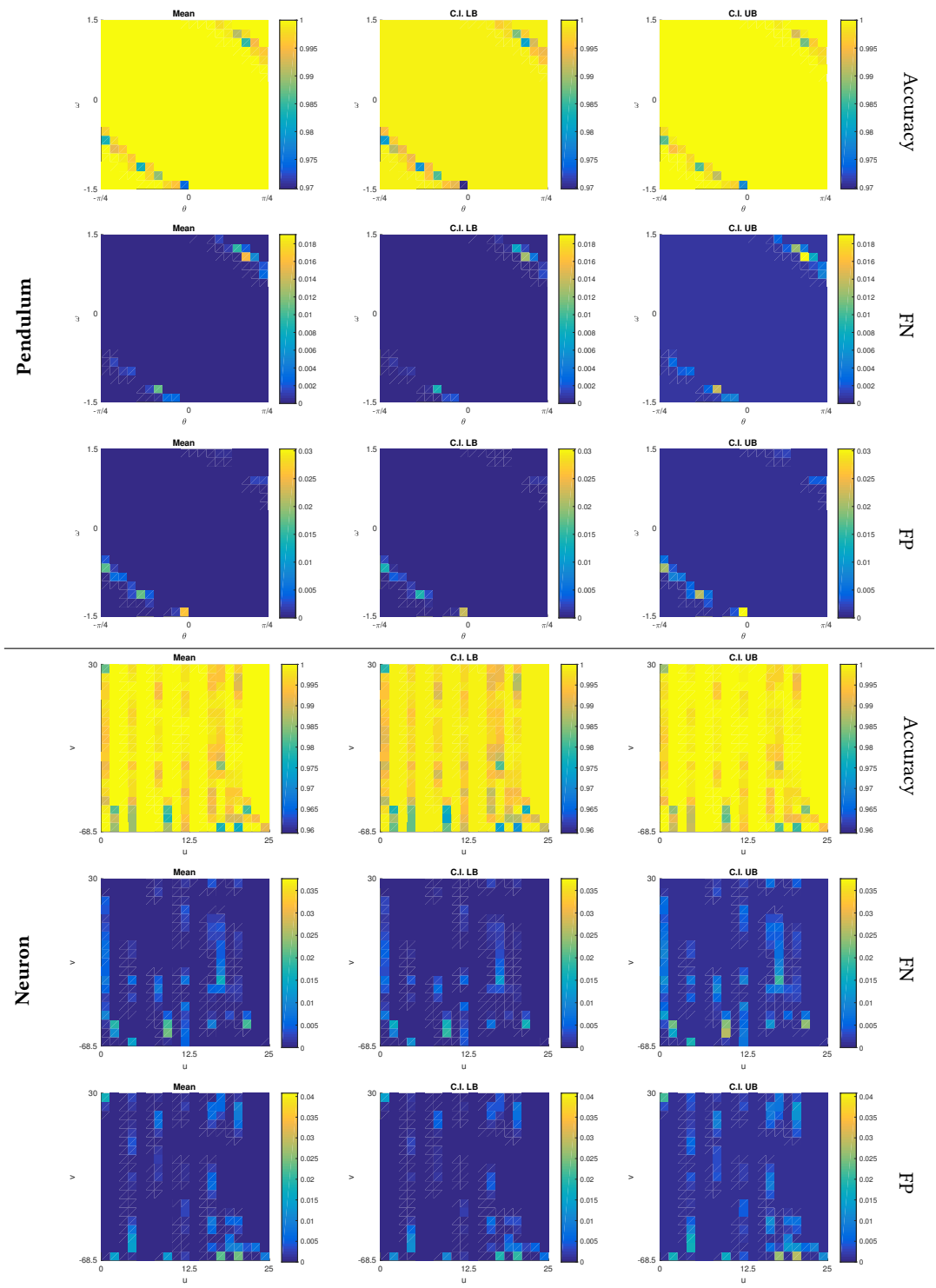

Figure 9: Evaluation of region-specific performance for the DNN classifier (trained with 20K samples) by generating test datasets over a 20×20 decomposition of the state space. First column: sample mean. Second and third columns: lower and upper bounds of 99% confidence interval.

### Neuron

|  | Acc≥ | | FN≤ | | FP≤ | |
|---|---|---|---|---|---|---|
|  | 99.5% | 99.8% | 0.5% | 0.2% | 0.5% | 0.2% |
| DNN-S | ✓ (6600) | ✗ (6900) | ✓ (2900) | ✓ (4500) | ✓ (2900) | ✓ (3400) |
| DNN-R | ✗ (5600) | ✗ (800) | ✓ (4800) | ✗ (1000) | ✓ (6000) | ✗ (1900) |
| SNN | ✗ (3500) | ✗ (1300) | ✓ (10200) | ✗ (3000) | ✗ (14000) | ✗ (2000) |
| SVM | ✗ (1400) | ✗ (300) | ✗ (9600) | ✗ (400) | ✗ (2700) | ✗ (400) |
| BDT | ✗ (1900) | ✗ (800) | ✓ (11900) | ✗ (5200) | ✓ (5400) | ✗ (900) |
| Ens1 | ✓ (3100) | ✓ (15500) | ✓ (3100) | ✓ (2300) | ✓ (2900) | ✓ (2900) |
| Ens2 | ✓ (5600) | ✗ (7100) | ✓ (3100) | ✓ (11700) | ✓ (3800) | ✓ (3400) |

### Pendulum

|  | Acc≥ | | FN≤ | | FP≤ | |
|---|---|---|---|---|---|---|
|  | 99.5% | 99.8% | 0.5% | 0.2% | 0.5% | 0.2% |
| DNN-S | ✓ (2500) | ✓ (3400) | ✓ (2300) | ✓ (2300) | ✓ (2500) | ✓ (2900) |
| DNN-R | ✓ (3300) | ✓ (3400) | ✓ (2300) | ✓ (2900) | ✓ (2500) | ✓ (2900) |
| SNN | ✓ (3100) | ✗ (3000) | ✓ (2900) | ✗ (2800) | ✓ (2500) | ✓ (3400) |
| SVM | ✓ (2900) | ✗ (5100) | ✓ (2900) | ✓ (5600) | ✓ (2300) | ✓ (2900) |
| BDT | ? (50000) | ✗ (1500) | ✓ (6200) | ✗ (5800) | ✓ (3800) | ✓ (2300) |
| Ens1 | ✓ (2300) | ✓ (3400) | ✓ (2300) | ✓ (2300) | ✓ (2700) | ✓ (2300) |
| Ens2 | ✓ (2500) | ✓ (2900) | ✓ (2300) | ✓ (3400) | ✓ (2500) | ✓ (2300) |

### Quadcopter

|  | Acc≥ | | FN≤ | | FP≤ | |
|---|---|---|---|---|---|---|
|  | 99.5% | 99.8% | 0.5% | 0.2% | 0.5% | 0.2% |
| DNN-S | ✓ (5200) | ✓ (3400) | ✓ (2500) | ✓ (2300) | ✓ (2500) | ✓ (5100) |
| DNN-R | ✓ (3300) | ✓ (16100) | ✓ (2500) | ✓ (15500) | ✓ (2300) | ✓ (5600) |
| SNN | ✓ (2700) | ✗ (13200) | ✓ (2700) | ✓ (2300) | ✓ (2500) | ✓ (3400) |
| SVM | ✗ (500) | ✗ (200) | ✗ (800) | ✗ (200) | ✗ (4800) | ✗ (800) |
| BDT | ? (50000) | ✗ (300) | ✓ (6200) | ✗ (4000) | ✓ (5000) | ✗ (9200) |
| Ens1 | ✓ (3100) | ✓ (6700) | ✓ (2300) | ✓ (2300) | ✓ (3100) | ✓ (3400) |
| Ens2 | ✓ (2700) | ✓ (9500) | ✓ (2700) | ✓ (2900) | ✓ (3100) | ✓ (3400) |

Table 5: A posteriori statistical guarantees for the classifiers (trained with 20K samples). Results were obtained using the sequential probability ratio test, with a maximum of 50,000 samples. In parenthesis are the number of samples required to reach the decision. A few results are undetermined (indicated with ?) after the 50,000 samples. Parameters of the test are $\alpha = \beta = 0.01$ and $\delta = 0.001$.

becomes larger. As a results, all previous FN samples are enclosed in this expanded region, i.e., they are correctly reclassified as positive. The enlarged positive region also means the adapted DNN is more conservative, producing more FPs as shown in Fig. 11 (b).

To make sure that the adapted DNN also generalizes to never-before-seen data, we tested it on another independent set of 10K samples. On this test dataset, the original DNN reports an overall accuracy of 99.78%, 9 FP samples, and 13 FN samples. On the other hand, the adapted DNN achieves an overall accuracy of 99.29%, 71 FP samples, and 0 FN sample. This result confirms that the adapted DNN is more conservative as expected.

### 5.5 Reducing False Negatives through Threshold Selection

We show through our case studies how accurate threshold selection (introduced in Section 3.5) can considerably reduce the FN rate. In Figure 12, we report the effect of different thresholds on accuracy, FN and FP for the DNN classifier trained with 20K samples and test dataset of 10K samples. As one can expect, the FN rate is monotonic increasing with respect to the threshold, while the FP rate is monotonic decreasing, with a huge loss of classification accuracy as the threshold approaches 0 or 1.

For the pendulum model, threshold selection is ineffective because the FN rate stays constant for $\theta \in [0.02, 0.5]$, and thus $\theta = 0.5$ remains the most adequate threshold as it does not penalize the FP rate. In contrast, for the neuron and quadcopter models, $\theta$ can be effectively tuned to improve the FN rate, inevitably but slightly sacrificing the FP rate and accuracy. After a simple visual inspection of the plots, for the quadcopter model, we can select $\theta = 0.34$, leading to a decrease of the FN rate from $7 \cdot 10^{-4}$ to $3 \cdot 10^{-4}$, and an overall accuracy loss of just 0.01%. For the neuron model, we can select $\theta = 0.37$, in this way reducing the FN rate from $9 \cdot 10^{-4}$ to $6 \cdot 10^{-4}$, with an accuracy loss of 0.07%.

A more systematic strategy consists in finding the threshold that minimizes the FN rate subject to prescribed bounds on the accuracy loss. If we allow for accuracy losses up to 0.1%, for the quadcopter model we can drastically reduce the FN rate to $10^{-4}$ ($\theta = 0.15$), and to $3 \cdot 10^{-4}$ for the neuron model ($\theta = 0.28$). If we further relax the bound on accuracy loss to 0.5%, we achieve an FN rate of 0 for the quadcopter model ($\theta = 0.05$), and of $10^{-4}$ for the neuron model ($\theta = 0.07$).

### 5.6 Time Bound Analysis

We assess the effect of different time bounds $T \in \mathbb{T}$ in the reachability formulas on the prediction accuracy of DNNs. This analysis is crucial to determine the ideal time bound to use for building a reliable classifier for model checking.

Intuition suggests that a long time bound leads to a more complicated decision border between positive and negative regions of the state space due to e.g., non-smooth dynamics, and as a consequence of degraded accuracy. On the other hand, prediction accuracy and its dependence on the time bound is highly model-dependent, since it is affected by properties of the dynamics like discontinuities and attractors. For instance, if a system stabilizes within time $T'$ starting from any state, then the decision border and prediction accuracy will remain constant for any reachability bound $T \geq T'$.

Our analysis, summarized in Figure 13, confirms that accuracy variations are model-dependent: for the quadcopter controller, we observe that accuracy is relatively constant up until $T = 16$, after which a steep decrease happens leading to approximately 2% drop at $T = 20$. In contrast, for the pendulum and spiking neuron case studies, accuracy is robust with respect to $T$, suggesting that the neural network can be employed for predicting reachability for longer time bounds.

## 6 RELATED WORK

We discuss related work on online model checking, simulation-based verification, machine-learning techniques in verification, formal analysis of neural networks and neural networks for control.

*Online model checking (OMC).* A number of approaches solve the OMC problem by providing safety guarantees up to a short time horizon, and by frequently updating these guarantees at runtime. In this category, Rinast et al. [58] presents an OMC technique for timed systems implemented in UPPAAL [9] based on graph-based techniques for reconstructing the model state space from the real-world system state. OMC for hybrid automata (HA) models is considered by [53], where estimation of a linear HA from observations and time-bounded verification are applied at runtime to a laser tracheotomy case study. The method of Sen et al. [61] for OMC of multi-threaded programs can predict safety violations from successful traces, by building a lattice of admissible executions consistent with event ordering. A control-theoretic approach is presented in [29], where future violations are predicted at runtime and prevented through control actions. Another class of methods for OMC (see e.g. [34, 59]) decompose the analysis into an offline phase, where the computationally expensive part of the analysis is carried out, and an online phase where the pre-computed results can be efficiently checked/refined using runtime information. This approach is similar to monitor synthesis and runtime verification via monitoring [51]. Calinescu et al. [14] propose a framework for self-adaptive software systems based on runtime quantitative verification of probabilistic models, while an incremental analysis technique suitable for OMC of MDPs is presented in [49].

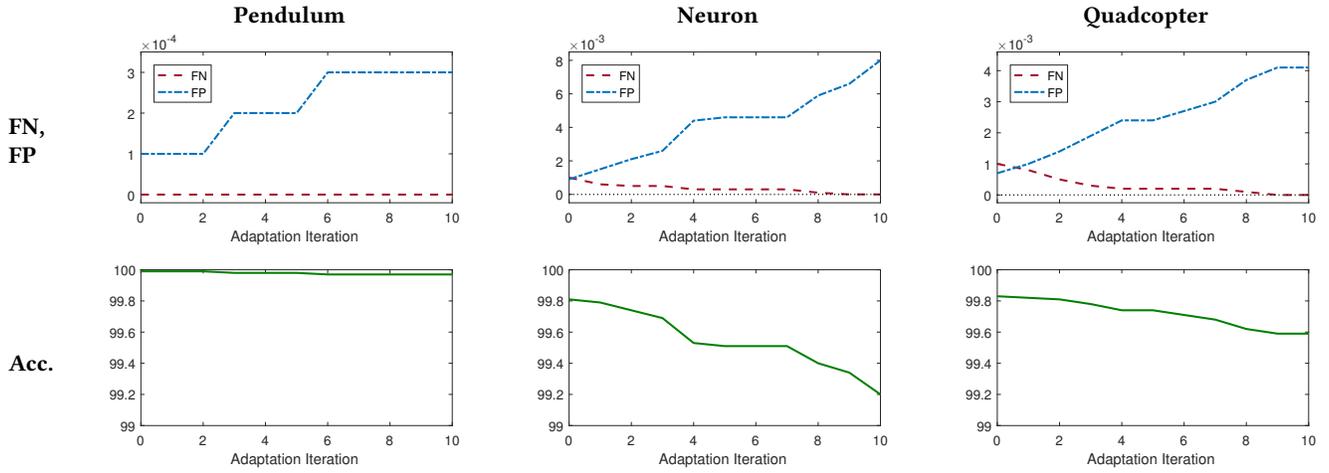

Figure 10: Impact of incremental adaptation on accuracy, false negatives and false positives, evaluated on the sigmoid-DNN classifiers trained with 20K samples and test dataset of 10K samples.

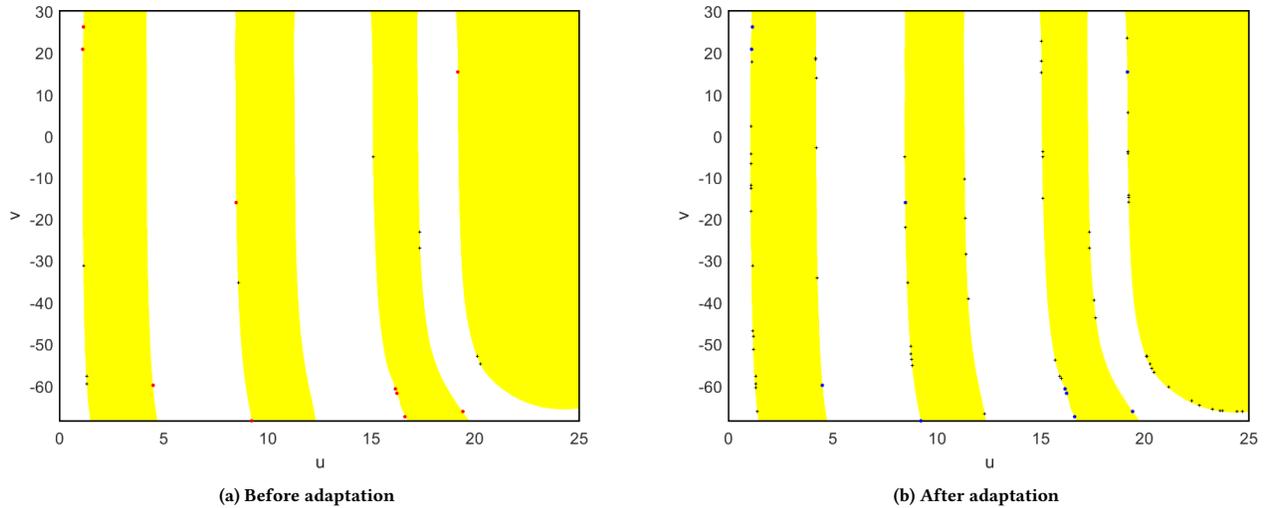

(a) Before adaptation

(b) After adaptation

Figure 11: Effects of adaptation on DNN-S trained with 20K samples for the spiking neuron case study. The white region is the predicted negative region. The yellow region is the predicted positive region. The crosses are FP samples. The red dots are FN samples. The blue dots are FN samples reclassified correctly as positive after adaptation.

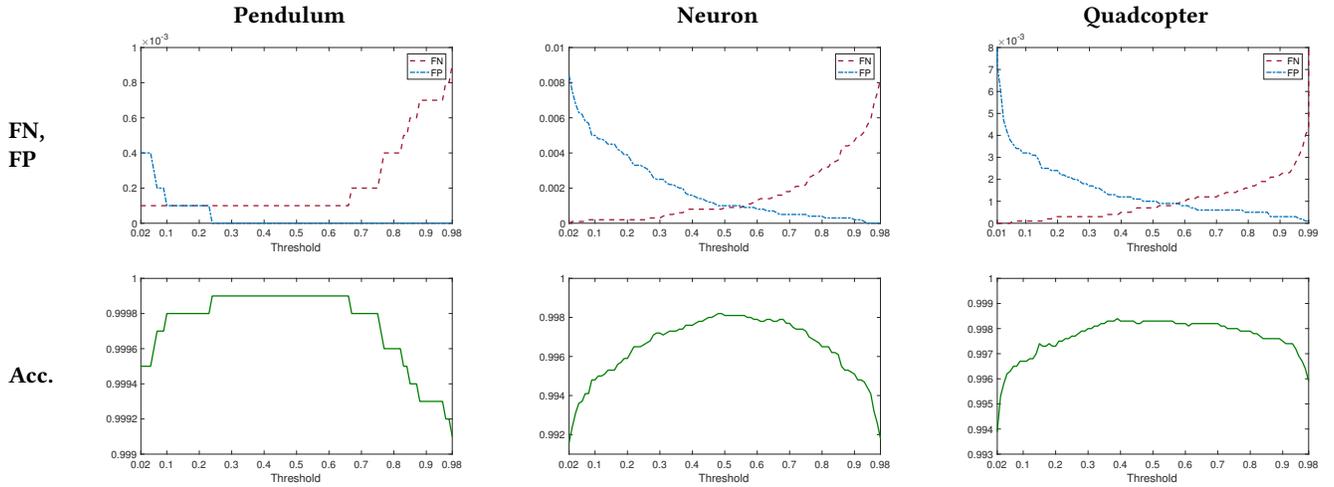

Figure 12: Impact of different classification thresholds (x-axis) on accuracy, false negatives and false positives, evaluated on the DNN classifier trained with 20K samples and test dataset of 10K samples. Note the different scales for the y-axis on the plots.

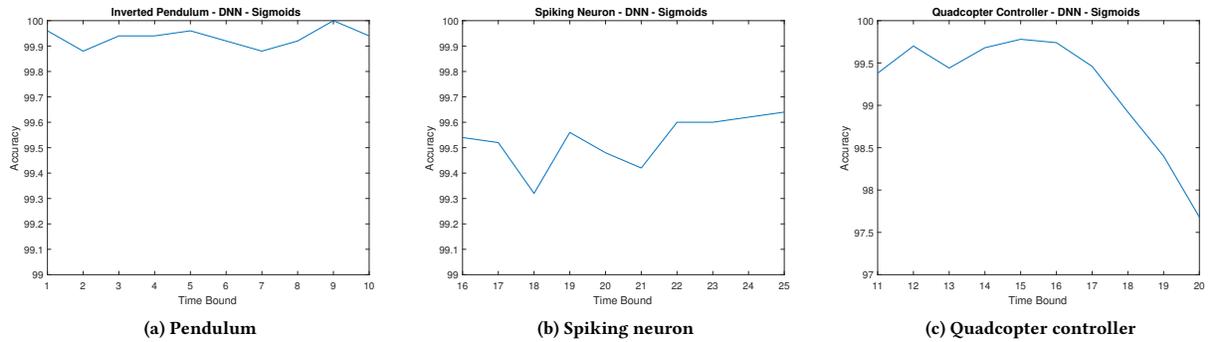

Figure 13: Time bound effects on the DNN prediction accuracy. DNNs were trained with 10,000 sample datasets and tested with 5,000 sample datasets.

Our NMC method also has an offline phase where we first learn from examples the approximate model checker, which can be queried at runtime in the online phase. None of the above approaches, however, employ machine-learning techniques for online model checking.

*Simulation-based verification.* Related work in this area centers around techniques for the rigorous analysis of hybrid and probabilistic systems from finitely many executions. Statistical model checking [50, 62, 71] relies on simulation and hypothesis testing to provide statistical guarantees (confidence intervals) on the probability that a given specification is satisfied by a probabilistic system. Other methods exist to estimate the satisfaction probability using Monte Carlo [36, 39] or Bayesian techniques [44, 74], as well as for stochastic hybrid systems [18, 32, 63, 67].

The approach of Donzé and others [23, 24] uses sensitivity analysis to compute an approximation of the set of reachable states for a given hybrid system, and hierarchical sampling to refine such approximations. A similar algorithm is developed in [27], which is based on "bloating" simulated trajectories of a hybrid system to obtain an over-approximation of the reachable set. Repeated sampling of system trajectories is also at the core of the S-Taliro tool [3] for the falsification of metric temporal logic (MTL) properties of nonlinear hybrid systems. This method exploits the robust (quantitative) semantics of MTL to drive the search towards traces with small robustness values, since negative robustness corresponds to violation of the property. The approach has been extended in [22] to generalize such counterexamples into larger falsifying regions (with probabilistic guarantees) using a combination of sampling and SMT solving. Bak et al. [4] build on the notion of super-position of linear systems to compute reachability of high-dimensional linear hybrid automata from simulations.

Similarly to these methods, our approach relies on sampling a finite number of executions, but we use these to train a classifier that provides (approximate) verdicts on time-bounded reachability. The above methods instead either focus on different problems (probabilistic model checking, falsification), or make restrictive assumptions on the dynamics, while we support arbitrary (black-box) deterministic dynamics: in [4], only linear hybrid systems are allowed; the work of [27] requires the user to specify discrepancy functions (a measure of trajectory convergence) which can be obtained automatically only for a limited class of systems [26]; in [24], an underlying ODE model is required to derive the variational equations describing sensitivity.

*Machine learning in verification.* Bortolussi and colleagues [11] apply Gaussian process (GP) regression and optimization [57] to infer the satisfaction function for continuous-time Markov chains, i.e., the function mapping model parameters to the corresponding satisfaction probability for a given property. Our work is similar in spirit but differs in two fundamental ways: 1) Our AMC problem is a classification problem due to the discrete (Boolean) reachability outcome; in contrast, in [11], the satisfaction function is continuous, thus yielding a regression problem. 2) In [11], model parameters constitute the input space and the time complexity of GP regression is strongly affected by the number of parameters. In contrast, NMC represents a function from the state-space to the Booleans, and its performance is not affected by the dimensionality of the system. GP-based techniques are also used for system design in [6, 7]. A solution based instead on genetic algorithms is presented in [13] for the robust design of probabilistic systems.

A problem related to verification is that of inferring temporal logic specifications from examples, solved in [8, 10, 64] by applying learning algorithms. Reinforcement learning [65] is commonly used in the analysis of Markov decision processes for policy learning in stochastic settings [2, 12], but is substantially different from the supervised learning techniques at the core of our work.

*Formal analysis of neural networks.* Motivated by the increasing number of applications of NNs in safety-critical tasks, in the last year the field of NN verification has been gaining great momentum, especially concerning the systematic derivation of adversarial examples, i.e. inputs able to "fool" the network inducing wrong predictions. We remark that, in our work, we seek to solve the opposite problem, i.e., that of training neural networks for predicting reachability.

One of the earliest works [56] introduces an abstraction-refinement method for safety verification and repair of NNs, where the abstractions are expressed in the theory of linear arithmetic and verified using an SMT solver. Scheibler and others [60] verify properties of a neural controller (based on sigmoid activation functions) for the inverted pendulum system and provide a direct encoding of the network in the theory of non linear reals, solved with the iSAT tool [30]. In [41], the authors present a method for finding adversarial inputs and robustness analysis for NNs for image classification, based on a layer-by-layer analysis and SMT techniques [19].

An SMT solver for the verification of ReLU feedforward networks is introduced by Katz and colleagues [45], which includes dedicated decision procedures (a modification of the simplex algorithm for linear programming) for this kind of networks. This approach is extended in [35] to automatically identify safe regions (i.e., immune to adversarial perturbations) through data-driven generation of candidate regions and formal verification of the candidates. The work of [31] solves the verification problem for NNs with piecewise-linear activation function based on combining SAT solving with linear programming and on linear approximations of the network behavior.

In [68], the synthesis of adversarial examples for image classification is reduced to a two-player turn-based stochastic game, where the first player seeks to find adversarial inputs by manipulating the features of the image, and the second player can be cooperative, adversarial, or random. Pei and others [55] take a different approach for the derivation of inputs inducing misclassification: given in input a set of networks trained for the same classification task, they employ gradient descent to find the inputs that maximize 1) the discrepancy among the predictions of these networks (indicating potential misclassification), and 2) a novel measure of network coverage.

Dutta et al. [28] tackle a different problem, that of computing rigorous and tight enclosures for the predictions of a ReLU NN over a convex input region (a form of "guaranteed range estimation"). The problem is solved with a combination of local search (gradient descent) and global optimization (mixed integer linear programming). The range estimation problem is also considered in [70], where a solution is proposed based on layer-by-layer sensitivity analysis. This problem is very similar to the estimation of prediction intervals [42, 46], where the enclosures are approximated by means of probabilistic and statistical techniques.

*Neural networks for control.* Since Hornik's seminal work [40] showing that feedforward neural networks with one input layers are universal approximators (i.e., able to approximate any continuous function), in the last two decades neural networks have been extensively applied to control problems. For a comprehensive study, we refer to the review [37]. Traditionally, neural networks are used for system identification, that is, to approximate the behavior of plants with unknown dynamics. The structure of such networks are inspired by autoregressive moving-average models, i.e., whose evolution is described by a non-linear function of sequences of past states and inputs. The identified network is then employed for controller design, typically as the prediction model in model-predictive control (MPC), or to train in turn a neural network-based controller. NNs have been also used in [33] to learn optimal switching policies among different controllers to ensure stability of the closed-loop system.

Widely applied to control problems in robotics, policy search is a reinforcement learning method that seeks to optimize parameters of a policy, described as state-dependent distributions of control actions [20, 47]. In [52], a guided policy search method is introduced for training neural network policies using an optimal control algorithm as a supervisor, thus making the problem one of supervised learning. The optimal control algorithm is typically an offline trajectory optimization procedure, or as in [72], policies are trained using an MPC controller, which makes the learned policy more robust to model errors and, compared to classical MPC, allows to circumvent the problem of state estimation.

# 7 CONCLUSIONS

We have shown how machine-learning techniques and specifically neural networks offer a very effective and highly efficient solution to the approximate model-checking problem for continuous and hybrid systems. To the best of our knowledge, we are the first to establish this link from machine learning to model checking.

There are many directions for future work to explore this link more broadly and to improve our current techniques. To improve accuracy, we plan to experiment with more sophisticated sampling techniques during training such as hierarchical sampling [25], rapidly-exploring random trees [17], or robustness-guided sampling [3], in order to thoroughly populate the training data with states that lie on the border of the positive and negative regions of the state space.

We also plan to examine a larger class of verification properties and extend NMC to systems with noisy and stochastic dynamics.